\documentclass[journal]{IEEEtran}
\ifCLASSINFOpdf
\else
\fi
\usepackage{lettrine}
\usepackage{amsfonts}
\usepackage{multirow}
\usepackage{commath}
\usepackage{mathtools}
\usepackage{makecell}
\usepackage{amsmath}
\usepackage[ruled,linesnumbered]{algorithm2e}
\newcommand{\etal}{\textit{et al.}}
\newcommand{\RN}[1]{%
  \textup{\uppercase\expandafter{\romannumeral#1}}%
}
\usepackage{ctable} 
\begin{document}
%
\title{Entropy-Based Uncertainty Calibration for Generalized Zero-Shot Learning}
%
%
%

\author{Zhi Chen, Zi Huang, Jingjing Li, Zheng Zhang
\thanks{Z. Chen, Z. Huang are with School of Information Technology \& Electrical Engineering, The University of Queensland, Brisbane, QLD 4072, Australia. (e-mails:uqzhichen@gmail.com, huang@itee.uq.edu.au).}

\thanks{J. Li is with the School of Computer Science and Engineering, University of Electronic Science and Technology of China, Chengdu 611731, China (e-mail: lijin117@yeah.net).}

\thanks{Z. Zhang is with with the Bio-Computing Research
Center, Shenzhen Graduate School, Harbin Institute of Technology, Shenzhen 518055, China (e-mail:  darrenzz219@gmail.com).}

}

\markboth{}%
{Shell \MakeLowercase{\textit{et al.}}: Bare Demo of IEEEtran.cls for IEEE Journals}

\maketitle

\begin{abstract}
Compared to conventional zero-shot learning (ZSL) where recognising unseen classes is the primary or only aim, the goal of generalized zero-shot learning (GZSL) is to recognise both seen and unseen classes. Most GZSL methods typically learn to synthesise visual representations from semantic information on the unseen classes. However, these types of models are prone to overfitting the seen classes, resulting in distribution overlap between the generated features of the seen and unseen classes. The overlapping region is filled with uncertainty as the model struggles to determine whether a test case from within the overlap is seen or unseen. Further, these generative methods suffer in scenarios with sparse training samples. The models struggle to learn the distribution of high dimensional visual features and, therefore, fail to capture the most discriminative inter-class features. To address these issues, in this paper, we propose a novel framework that leverages dual variational autoencoders with a triplet loss to learn discriminative latent features and applies the entropy-based calibration to minimize the uncertainty in the overlapped area between the seen and unseen classes. Specifically, the dual generative model with the triplet loss synthesises inter-class discriminative latent features that can be mapped from either visual or semantic space. To calibrate the uncertainty for seen classes, we calculate the entropy over the softmax probability distribution from a general classifier. With this approach, recognising the seen samples within the seen classes is relatively straightforward, and there is less risk that a seen sample will be misclassified into an unseen class in the overlapped region. Extensive experiments on six benchmark datasets demonstrate that the proposed method outperforms state-of-the-art approaches.
\end{abstract}

\begin{IEEEkeywords}
Generalized zero shot learning, image classification, transfer learning, triplet network
\end{IEEEkeywords}

%
\IEEEpeerreviewmaketitle

\section{Introduction}
\IEEEPARstart{O}{bject} recognition has seen remarkable advancements since the resurgence of deep convolutional neural networks \cite{krizhevsky2012imagenet, he2016deep}. However, its alluring performance is expensive, with a typical recognition model requiring an inordinate amount of labelled data for training. Without high-quality annotated data, supervised learning breaks down with no way to ensure that a model will be able to predict, classify, or otherwise analyze the phenomenon of interest with any accuracy \cite{simonyan2014very, luo2020progressive,wang2020prototype,luo2020adversarial,luo2019learning}. Beyond the obvious problems with constructing a robust model, crossing the desert of available data to scale up recognition systems is an arduous, if not impossible, journey. 

As a method for overcoming this challenge, zero-shot learning (ZSL) \cite{shen2019scalable,long2018pseudo,yang2016zero,luo2017zero,luo2017zero} has been hailed as something of a caravan. With ZSL, models are trained on a small amount of supervised data, i.e., seen classes, where they learn to “mix and match” the features they know to classify unseen objects. This learning paradigm is a subfield of transfer learning, akin to taking knowledge learned from the source domain and using it to complete a task in the target domain \cite{wang2019survey}. Generalized zero-shot learnning (GZSL) \cite{chao2016empirical,rahman2018unified,li2019alleviating,li2019zero,chen2020rethinking} is a more practical but also more challenging direction of this research that aims to recognise objects in the 
target domain, i.e., unseen classes while still being able to recognise objects in the source domain, i.e., seen classes. 

\begin{figure}[t]
\centering
\includegraphics[width=0.95\columnwidth]{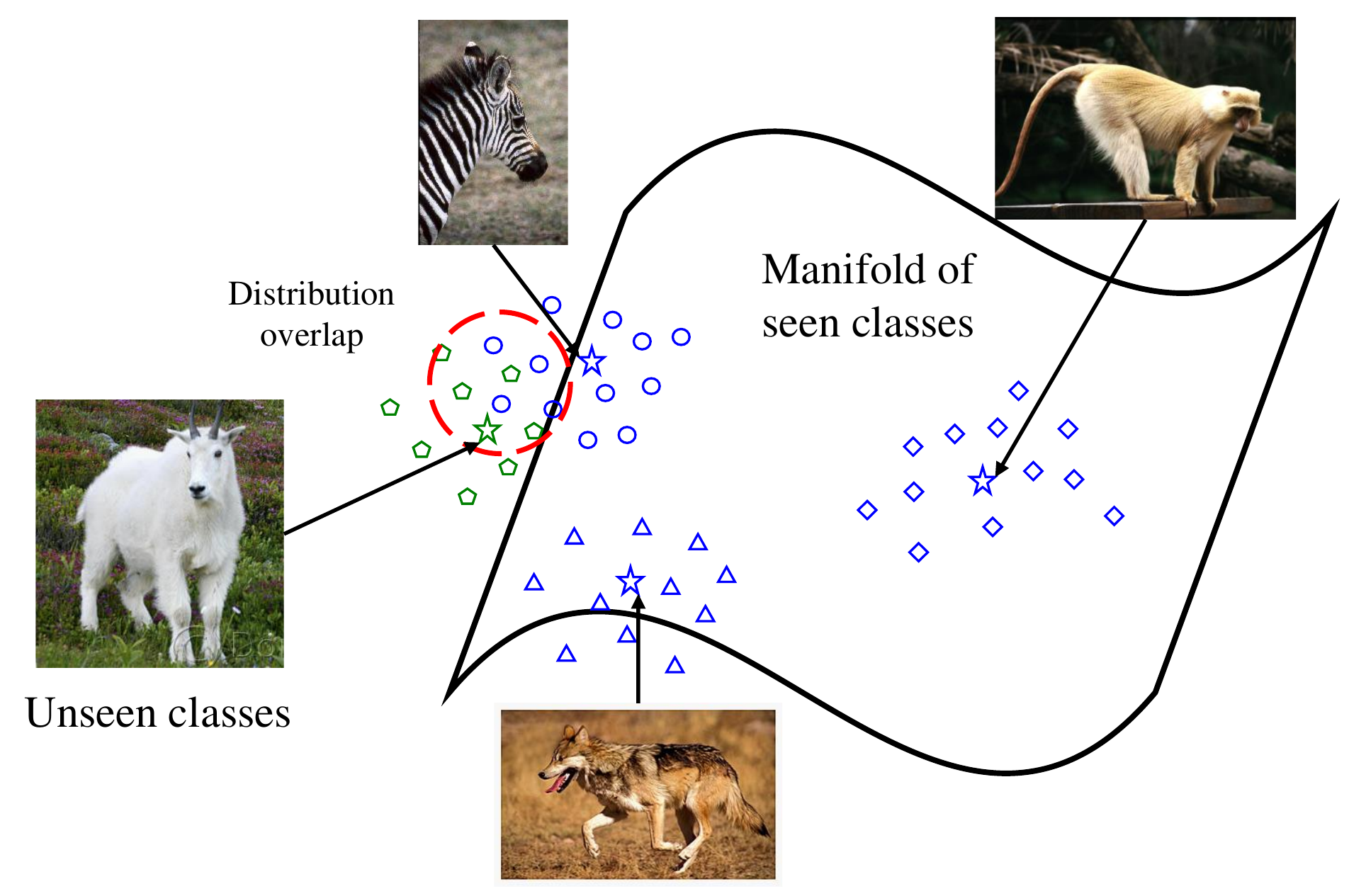} 
\caption{
Distribution overlaps between the seen and unseen classes. GZSL methods tend to overfit the seen classes, which covers a wider range than the underlying real distribution. This unexpected overlap with the unseen classes leads to reduced prediction accuracy with the seen classes.
}
\label{fig1}
\end{figure}

Some state-of-the-art GZSL approaches \cite{zhu2018generative, kumar2018generalized, huang2019generative, kodirov2017semantic, chen2020canzsl, li2020learning} use generative models, such as generative adversarial nets (GANs) \cite{goodfellow2014generative}, variational autoencoders (VAEs) \cite{VAE}, or different variants of hybrid GAN/VAEs, to find an alignment between class-level semantic descriptions and visual representations. As such, the zero-shot learning problem becomes one of a traditional supervised classification task. However, there are two main limitations still to be overcome with existing GZSL methods. These are \textbf{sparse training samples} \cite{zhu2018generative} and \textbf{distribution overlaps}. The sparsity of training samples are common in most image datasets, the distribution of high dimensional visual features are thus hard to learn. As such, the generative models fail to capture the most discriminative inter-class features.

Distribution overlap refers to the potential outcome that a generative model trained only on seen classes becomes overfit to those classes. Figure 1 illustrates how, in this circumstance, the distribution of the synthesised features for the unseen classes can partially, or even fully overlap, with the seen class distribution. The result is uncertainty about whether a test case is seen or unseen and, ultimately, a model with sub-par performance on seen classes. In our experiments, for instance, a simple softmax classifier solely trained on the seen classes was able to achieve more than 90\% accuracy on the aPaY dataset. But when unseen samples were included, the seen class accuracy dropped to 51.8\%.

To address these issues, we developed a discriminative latent feature generation model with an entropy-based uncertainty calibration technique for GZSL. The generative model contains two variational autoencoders (VAEs) with a triplet loss that learn inter-class discriminative latent features. The encoders in the two VAEs synthesise latent feature embeddings; one is dedicated to the visual space, the other to the semantic space. The latent spaces for the two VAEs are constrained to be shared. Hence, both are able to distil the visual and semantic information. 
The triplet loss helps to avoid the latent features of each class collapsing into very small clusters by training the generation model to distinguish between similar and dissimilar pairs of examples. This approach minimises the distances between generated latent features from the same class and vice versa for samples in different classes. 
The remaining two decoders reconstruct the visual and semantic representations from the latent features. An uncertainty calibration is performed during the classification stage, where the latent features of the seen classes are embedded via the visual feature encoder, while the latent features of the unseen classes are generated with the semantic encoder. A general classifier is then constructed over both the seen and unseen classes to classify the latent features. During the testing phase, the general classifier calculates the probability entropy over the seen classes from a softmax output. The probability entropy is a measure of the uncertainty as to whether the sample is seen or unseen. Samples are deemed seen if the entropy value falls below a tuned threshold, and a simple visual classifier is then trained on all the definitively seen samples. The rest of the uncertain test samples are fed into the general classifier to make the final predictions. 

In summary, the contributions of this paper are therefore as follows:
\begin{itemize}
    \item 
    We propose a novel technique for GZSL that exploits an entropy-based uncertainty calibration (EUC) to alleviate issues with distribution overlaps, by training cascade classifiers. Unlike conventional probability calibration techniques that tune a softmax temperature, we propose probability entropy as a measure of confidence that a sample belongs to a seen class.
    \item 
    A novel generative metric learning (GML) paradigm is presented that leverages dual variational autoencoders with a triplet loss to synthesise latent features. The triplet loss helps to capture the inter-class discrimination in scenarios with sparse training samples.
    \item 
    Extensive experiments on six benchmark datasets demonstrate the superior performance of the proposed method in both conventional and generalized ZSL against the current state-of-the-art methods.
\end{itemize}

\section{Related Work}
\subsection{Generative Models for ZSL}
Recent state-of-the-art approaches to GZSL with generative models have achieved promising performance. Generative models can synthesise an unlimited number of fake features from side information on the novel classes, e.g., semantic attributes. With these synthesised features, ZSL problems become a relatively straightforward supervised classification task. The two most commonly used generative models are generative adversarial networks (GANs) \cite{goodfellow2014generative} and variational autoencoders (VAEs) \cite{VAE}. Often, both models are jointly used to form generative architectures for ZSL tasks. GAZSL \cite{zhu2018generative} leverages Wasserstein GANs (WGAN) \cite{arjovsky2017wasserstein} to synthesise vivid visual features. Currently, GAZSL models are the state-of-the-art for ZSL tasks with noisy text. CANZSL \cite{chen2019canzsl} is a cycle-consistent generative framework for solving domain shift problems and enhancing denoising operations on natural language input. SAE \cite{kodirov2017semantic} is a semantic autoencoder that learns mappings between the semantic space and the visual space for the encoder to use to translate visual features into the semantic space. The decoder can then synthesise unlimited visual features from the semantic information. GDAN \cite{huang2019generative} takes a different approach with a novel structure consisting of a conditional variational autoencoder (CVAE) and a GAN that generate various visual features from class feature embeddings as input. SE-ZSL \cite{kumar2018generalized} adopts an autoencoder followed by an attribute regressor to train a model with three alignments: visual-to-attribute, attribute-to-visual and visual-to-attribute. The latter two feedback alignments help to improve the effectiveness of the decoder, which works as a visual feature generator.

However, in these methods, the feature generation between the semantic and visual spaces is constrained by asymmetric information. For example, suppose that the wing colour in a bird image is not annotated, so the visual features generated from the semantic labels do not include this information. However, the visual feature extractor can and does extract the wing colour. The two visual representations are now asymmetric, so training a classifier from the generated visual features but testing with extracted visual features may result in sub-optimal performance.

With this motivation, our framework incorporates two variational autoencoders with triplet metric regulation to learn latent features that best discriminate between the visual and semantic spaces.

\subsection{Uncertainty Mitigation}
One strategy for mitigating the distribution overlap issue is to simply distinguish between seen and unseen classes \cite{dong2018learning, gidaris2018dynamic, li2019leveraging}. This strategy is a challenging yet rewarding approach in transfer learning tasks, such as GZSL, few-shot learning and domain adaptation. Gidaris and Komodakis \cite{gidaris2018dynamic} proposed a few-shot classification weight generator in a recent work to differentiate seen from unseen classes. LisGAN \cite{li2019leveraging} presented a technique for GZSL to leverage unseen samples given a high level of confidence that the samples had been correctly classified. The high-confidence unseen samples are then regarded as references for future classification. OSBP \cite{saito2018open}, an approach to domain adaptation, trains the feature generator on representations so as to separate unknown target samples from known ones.

Another strategy to balance performance between seen and unseen classes is to apply a softmax probability calibration \cite{liu2018generalized, chao2016empirical, das2019zero}. From a realization that performance with unseen class in GZSL was unsatisfactory, Chao \textit{et al.} \cite{chao2016empirical} proposed a shifted-calibration approach to improve performance. DCN \cite{liu2018generalized} incorporates temperature calibration to enable simultaneous calibration of deep networks based on confidence in the source classes and uncertainty in the target classes. Tuning the temperature for softmax layer is conducted through knowledge distillation \cite{hinton2015distilling}, which produces softer probability distributions over the classes. Temperature calibration in DCN, however, has another exciting purpose, which is to mitigate the uncertainty for unseen classes in GZSL. RAC \cite{das2019zero} involves scaled multiplicative calibration of the classification scores of the seen classes, which changes the effective variance of the seen categories.

Unlike the methods above that directly calibrate the probability distribution of the seen classes, our framework is based on a probability entropy over the seen classes as a measure of the confidence that a class is seen.

\subsection{Metric Learning in ZSL}
\begin{figure*}[t]
\centering
\includegraphics[width=0.95\textwidth]{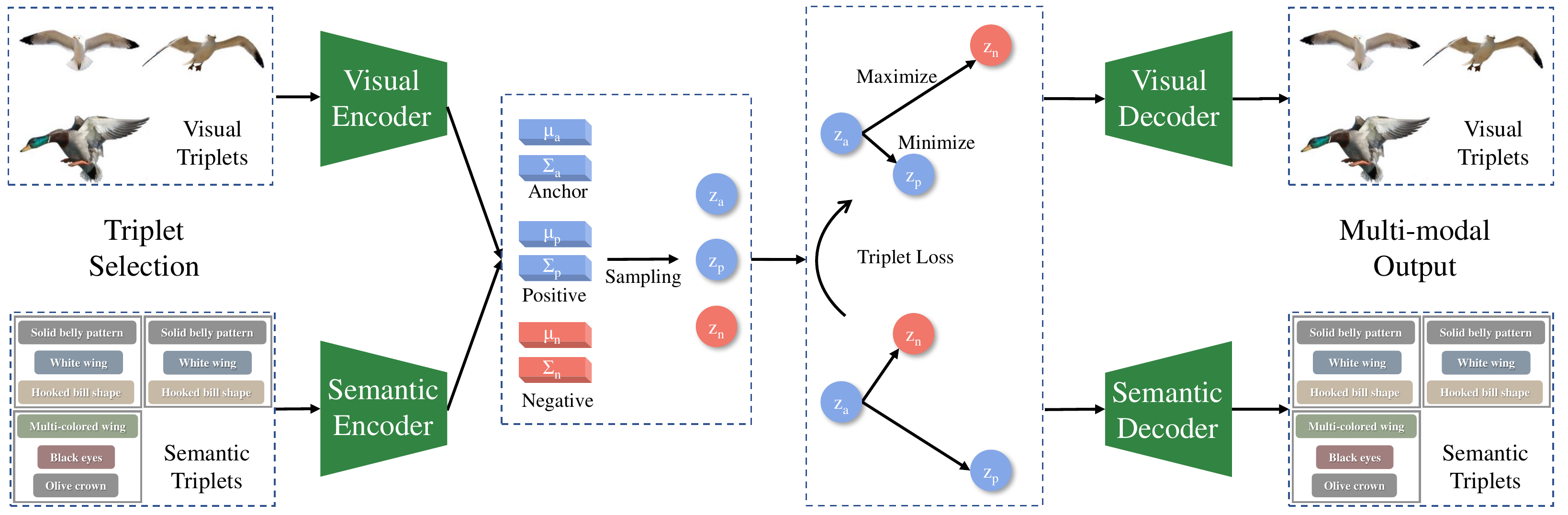} 

\caption{
The proposed GML approach for GZSL. The triplet input consists of an anchor, a positive sample from the target class and a negative sample from one of the different classes. The visual triplet is then fed into the encoder of the visual VAE and, likewise, with the semantic triplets and the encoder of the semantic VAE. The latent vectors are then sampled from the synthesized mean and covariance vectors. Further, a triplet loss is applied to those latent vectors before decoding them into visual and semantic triplets to calculate reconstruction losses.
}
\label{fig2}

\end{figure*}
When used in conjunction with distance or similarity-based techniques, such as nearest-neighbour approaches, the promise of metric learning methods becomes very clear. Learning an effective distance metric among visual pairs is a significant paradigm in visual analysis, with a number of recent studies in this area \cite{ge2018deep,duan2018deep,schroff2015facenet}. The two commonly used networks for metric learning are the Siamese network and the triplet network. Inspired by DeepRanking \cite{wang2014learning}, triplet networks \cite{hoffer2015deep} have been shown to learn better representations than Siamese networks. Hence, triplet networks have become widely used in distance metric learning tasks, like image retrieval and face verification \cite{ge2018deep,duan2018deep,schroff2015facenet}. FaceNet \cite{schroff2015facenet} uses triplets of roughly aligned matching and non-matching face pairs selected with an online triplet mining approach. The result is state-of-the-art face recognition performance. DAML \cite{duan2018deep} generates hard negative samples from observed samples in an adversarial learning manner for both image retrieval and clustering. Weifeng Ge \cite{ge2018deep} developed a novel hierarchical triplet loss (HTL) for image retrieval tasks that collects informative training samples with the guide of a global class-level hierarchical tree with favourable performance over vanilla triplet loss. We have also taken advantage of triplet loss in metric learning and incorporated it into our generative model to synthesise discriminative latent space features.

Metric learning has also been seen in recent approaches to ZSL. For example, GAZSL \cite{zhu2018generative} involves a technique called visual pivot to force the generated visual features to approximate the cluster centres in each class, which means nearest-neighbour approaches perform quite well. However, the fixed visual pivots are only able to pull samples with the same label into nearby points; they lack the ability to push samples of different classes apart. GDAN \cite{huang2019generative}  uses a regressor in additional to a generative model to learn a flexible distance with which to evaluate the closeness of the visual features and class embeddings. Experiments show, however, that the model performs extremely well with seen classes, but unseen objects are prone to bias toward seen categories. Conversely, our metric learning regulation is explicitly applied to the latent feature space to achieve good discrimination.

\section{Methodology}
In this section, we first introduce our problem formulation and then detail the dual generative framework that produces the latent features. The section concludes with a discussion on the triplet loss and the uncertainty calibration mechanism for predicting the labels of the test samples.

\subsection{Problem Definition}
Let $\mathcal{T}$ be the set of all possible triplets $ \{\tau_{a}, \tau_{p}, \tau_{n}\}$ from the seen classes $S$, where $ \tau_{a}$ and $\tau_{p}$ are from the same category, and $\tau_{n}$ is a negative sample randomly chosen from a different category. 

Each $\tau$ contains a triplet $(x,s,y)$, where $x$ denotes the visual features extracted from an image, $s$ represents the corresponding semantic attributes and $y$ is the associated one-hot class label, respectively. During the evaluation phase, visual features $x^{u}$ and the corresponding semantic representations $s^{u}$ are drawn from the unseen classes $U$. In ZSL, the aim is to predict the class label $y^{u}$. By comparison, the aim with GZSL is to predict the class label for the image $x$ from both $S$ and $U$.

\subsection{Basic Generative Module}
The basic module in our framework is a variational autoencoder (VAE) \cite{VAE} that includes an encoder, a decoder and an objective function. The VAE models the probabilistic generation of data, and the encoder and decoder networks are probabilistic. Specifically, we can compute the mean $\mu$ and the diagonal covariance $\Sigma$ of the latent conditional distribution $z$ from $x$ with the encoder Q. The latent data $z$ is then sampled from $\mathcal{N}(\mu_{z|x},\Sigma_{z|x})$. Further, the reconstructed or generated data is synthesised by the decoder $P(x|z)$. The objective function of a VAE has a tractable lower bound. Therefore, we can optimize the gradient of this bound with the following formula:
\begin{equation}
  \mathbb{E}_{Q_{\phi}(z|x)}[logP_{\theta}(x|z)] - D_{KL}[Q_{\phi}(z|x)||P_{\theta}(z)],
\end{equation}
where the first term is the reconstruction loss of the input data, and the second term is the Kullback-Leibler (KL) divergence between the encoder model and the simple Gaussian prior. Here, the functions $Q_{\phi}$ and $P_{\theta}$ are the encoder and decoder networks with the corresponding weights $\phi$ and $\theta$, respectively.

\subsection{Dual Generative Model}
As shown in Figure 2, the synthesised features are regularised in the latent space according to the labels by applying a triplet loss. The latent space is formed between the visual and semantic spaces through a dual VAE framework, which includes a visual variational autoencoder $v$-VAE and a semantic variational autoencoder $s$-VAE. In terms of the implementation of the dual VAEs, we followed the CADA-VAE approach set out in \cite{schonfeld2019generalized}.

The conventional role of VAEs as a generative model leverages the decoder as the generator. By contrast, both the encoders in our dual generative framework work as generators to synthesise embedding vectors in the latent space, while the decoders translate the latent vectors back into visual or semantic features. In other words, the VAE has a cycle structure, providing supervision information to improve the generation ability of the encoders.

The VAE loss for $v$-VAE and $s$-VAE can be formulated as:
\begin{equation}
\begin{aligned}
  \mathcal{L}_{v\mathrm{\text{-}VAE}} = \mathbb{E}_{Q_{v}(z_{v}|x)}[logP_{v}(x|z_{v})] \\ - \beta_{1} D_{KL}[Q_{v}(z_{v}|x)||P_{v}(z_{v})],
\end{aligned}
\end{equation}
\begin{equation}
\begin{aligned}
  \mathcal{L}_{s\mathrm{\text{-}VAE}} = \mathbb{E}_{Q_{s}(z_{s}|s)}[logP_{s}(s|z_{s})]  ~\\ - \beta_{2} D_{KL}[Q_{s}(z_{s}|s)||P_{s}(z_{s})],
\end{aligned}
\end{equation}
where $Q_{v}$, $Q_{s}$ denote the encoder networks in $v$-VAE and $s$-VAE respectively, and $P_{v}$, $P_{s}$ represents decoder networks. The first terms are the reconstruction loss and the second ones are the KL divergence. $\beta_{1}$ and $\beta_{2}$ are the coefficients of KL divergence for the two VAEs, respectively. $z_{v}$ and $z_{s}$ are the latent vectors synthesised from the visual and semantic spaces, respectively. In more detail, the Gaussian distributions $\mathcal{N}_{v}(\mu_{x},\Sigma_{x})$, $\mathcal{N}_{s}(\mu_{s},\Sigma_{s})$ are the direct output of $Q_{v}$ and $Q_{s}$. $z_{v}$ and $z_{s}$ sampled with a reparametrization trick \cite{VAE} from the two Gaussian distributions.

To make the latent space and intermediate region between the visual and semantic spaces, we use a multi-distribution loss to allow the two Gaussian distributions to approximate each other. In our experiments, we found that a Wasserstein distance gave better performance than other distance functions, such as mean squared error (MSE). Hence, the cross-modal loss is formulated as follows:
\begin{flalign}
\mathbf{W}(\mathcal{N}_{v}, \mathcal{N}_{s})^{2} &=   \   \lVert \mu_{v}-\mu_{s}\rVert^{2}_{2}   \\ &+  trace(\Sigma_{v} + \Sigma_{s} - 2 (\Sigma_{s}^{1/2} \Sigma_{v} \nonumber \Sigma_{s}^{1/2})^{1/2}).
\end{flalign}
Further, the latent vectors  $z_{v}$ synthesised from the visual space can be decoded into semantic embeddings $\hat{s} \leftarrow P_{s}(z_{v})$ and vice versa for the latent vectors $z_{s}$ encoded from the semantic space, $\hat{v} \leftarrow P_{v}(z_{s})$. The objective function for optimizing the multi-modal reconstruction is given below:

\begin{equation}
  \mathcal{L}_{mul-recon} = \lVert \hat{v} - v\rVert_{1} + \lVert \hat{s} - s\rVert_{1}, 
\end{equation}
where we use an L1 norm because is it insensitive to outliers. With an L2 norm, too many outliers in the dataset will degrade performance.

\subsection{Triplet Regularization}
To regularize the latent features for optimal discriminativeness, we have incorporated a triplet loss into the method in addition to the above constraints. As shown in Figure 2, a triplet loss is enforced in the latent space to ensure a small distance between all objects of the same category and a large pairwise distance between objects different categories. Training is performed in batches, with selected triplet batches, including an anchor batch, a positive batch and a negative batch. The samples in the positive batch have the same corresponding labels as the anchor batch, whereas the negative batch samples all have different random labels. Formally, the triplet loss can be written as follows:
\begin{flalign}
  \mathcal{L}_{v\text{-}trip} = max(\mathcal{D}, 0), 
\end{flalign}
\begin{flalign}
  \mathcal{D} = \lVert Q_{v}(x_{a})-Q_{v}(x_{p})\rVert^{2} - \lVert Q_{v}(x_{a})-Q_{v}(x_{n})\rVert^2 + \alpha,
\end{flalign}
where $x_{a}$, $x_{p}$ and $x_{n}$ are the anchor images, the positive samples and the negative samples, respectively. $\alpha$ represents the margin between the positive and negative pairs, which varies from dataset to dataset. Similar triplet losses $\mathcal{L}_{s\text{-} triplet}$ with the same $\alpha$ are also applied to the latent features $z_{s}$ generated from the semantic embeddings.

To further improve multi-modal reconstruction and regularize the margin between classes, we propose the following multi-modal triplet objective function:
\begin{flalign}
  \mathcal{L}_{mul\text{-}trip} &= {\sum^{M}_{i}\sum^{M}_{j}\sum^{M}_{m}}_{!(i=j=m)} max(\mathcal{D}_{i,j,m} , 0 \ ), \\
\end{flalign}
\begin{flalign}
  \mathcal{D}_{i,j,m} = \lVert Q_{i}(i_{a})-Q_{j}(j_{p})\rVert^{2}  - ||Q_{i}(i_{a})-Q_{m}(m_{n})||^2 + \alpha,
\end{flalign}
where i, j and m represent the modality M, which can be either visual or semantic. Note that i, j and m cannot be the same modality. This objective function specifies six more triplet loss terms in the latent space.

The superiority of the soft margin in triplet loss over fixed visual centres (aka. visual pivots) as proposed in GAZSL\cite{zhu2018generative} is worth discussing. A visual centre is simply calculated by the mean value of all image features in the same class, and the synthesised visual features from the same class are pushed towards the corresponding visual centre. However, while optimizing the distance between the visual data points, the model may converge into a situation where the samples from different categories overlap. The triplet loss in our framework not only pulls the samples of the same category towards each other, it also pushes negative category objects backwards.

The overall objective function for the proposed approach GML is as follows:
\begin{flalign}
  \mathcal{L}_{GML} &=  \mathcal{L}_{v\mathrm{\text{-} VAE}} +  \mathcal{L}_{s\mathrm{\text{-} VAE}} + \lambda \mathbf{W}(\mathcal{N}_{v}, \mathcal{N}_{s})^{2} ~\\ &+   \mathcal{L}_{mul-recon} +  \mathcal{L}_{v\text{-} triplet} +  \mathcal{L}_{mul\text{-} trip} , \nonumber
\end{flalign}
where $\lambda$ denotes the weight of Wasserstein distance.

\subsection{Predicting with Uncertainty Calibration}
To mitigate catastrophic distribution overlap issues, we further develop a novel entropy-based uncertainty calibration technique to predict the labels of test samples. Once the dual VAE model has been trained, arbitrary instances can be generated in the latent space based on semantic embeddings of the classes. In GZSL, a softmax classifier is trained over both the seen and unseen classes with 200 latent representations for each seen class. For datasets that do not have enough samples per class, the available instances are simply repeated until 200 samples have been trained. For example, if only 50 samples were available, the training set would consist of four duplicates of the 50 samples. Plus, there are fewer unseen categories in the dataset than seen ones. To avoid an imbalance of knowledge to be learned between the seen and unseen classes, more samples of the unseen classes can be synthesised, say 400. Note that the general classifier training data of seen classes are synthesised by the visual encoder $Q_{v}$, whereas the latent features for the novel classes are all synthesised from semantic embeddings.

 There are three networks involved in predicting test samples: a general classifier $f$, a seen classifier $g$ and a visual encoder $Q_{V}$, as shown in Figure 3. Specifically, the general classifier is a softmax classifier that recognizes both seen and unseen samples. Latent features from seen classes and unseen classes need to be provided to train this general classifier. The seen latent features are mapped from the visual samples in the training set by the visual encoder $Q_{v}$. Unlimited novel latent features for the unseen classes can be synthesised with the trained semantic encoder $Q_{s}$ given class embeddings. To predict the labels in conventional GZSL, latent representations are first synthesised by the visual encoder given test the visual feature of the sample. And, once the general classifier is trained, recognizing which class a sample belongs is a straightforward task. When classifying objects into seen classes, almost all unseen objects will have high entropy, whereas approximately half of the seen samples will have lower entropy than the unseen ones. Therefore, entropy is calculated with unnormalized log probabilities from the softmax output. The insight here is that our latent feature generator is trained on the seen classes so whether an object is seen or unseen is more certain as determined by the probability entropy. 

Nevertheless, this approach does not prevent overlaps between the seen domain and the unseen domain in the same latent space and, in turn, performance degradation. Therefore, we incorporated an entropy threshold to judge whether an object is likely to be seen; thus, reducing the risk of the general classifier classifying an object into the wrong unseen class. After excluding the uncertain objects, a simple softmax classifier is trained in the seen domain to recognize the low entropy objects. Note that, instead of latent representations, this classifier is trained with visual features, which result in greater than 90\% accuracy with seen objects in the testing stage. However, a side effect of the proposed uncertainty calibrator is that performance with unseen classes can degrade slightly since a small portion of novel class samples are improperly classified into seen classes. Nevertheless, this tiny sacrifice of accuracy in the unseen domain is more than offset by the substantial improvement in accuracy in the seen domain. A more detailed comparison of merit with and without the uncertainty calibrator is provided in the ablation study.

\begin{figure}[t]
\centering
\includegraphics[width=1.0\columnwidth]{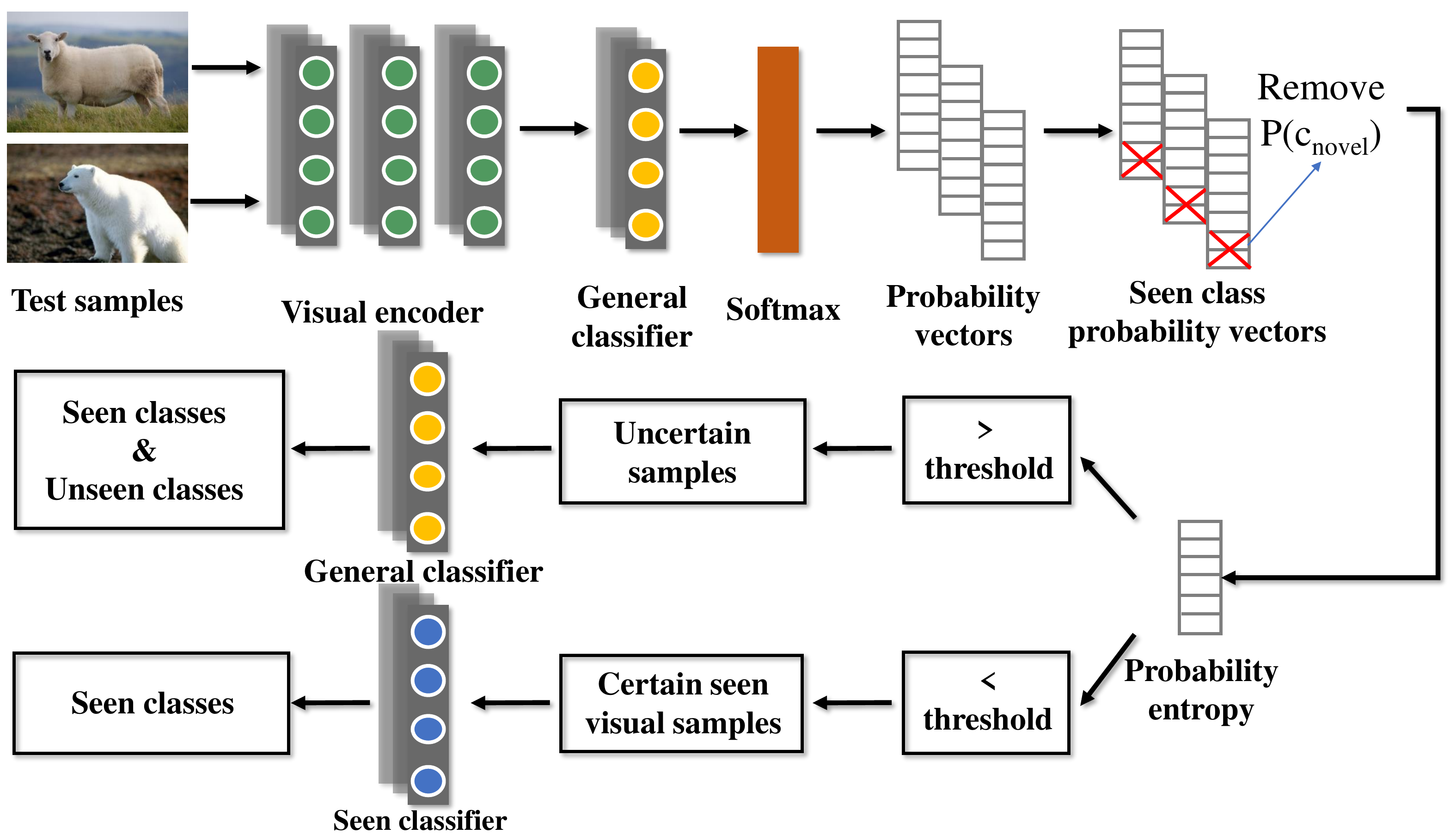} 

\caption{The flow diagram of predicting test samples with uncertainty calibrator mechanism. Samples with low probability entropy are regarded as source domain objects, and then classified within seen classes. The rest of samples are fed into a conventional GZSL supervised classifier.}
\label{fig1}

\end{figure}

\begin {table}[b]
\caption {Statistics of the six zero-shot learning datasets. }

\begin{center}
\scalebox{1.}{
\begin{tabular}{l|cccccc}
\hline
Dataset      & CUB    & aPaY     & AWA1   & AWA2   & SUN   & FLO    \\ \hline
\#Attributes & 312     & 64     & 85     & 85     & 102    & 1024   \\
\#Source     & 150     & 20     & 40     & 40     & 645    & 82     \\
\#Target     & 50      & 12     & 10     & 10     & 72     & 20     \\
\#Images     & 11,788  & 18,627 & 30,475 & 37,322 & 14,340 & 8,189  \\ \hline
\end{tabular}}
\end{center}
\end{table} 

\begin {table*}[t]
\caption { GAZSL accuracy (\%) on six datasets. U, S and H represent unseen, seen and Harmonic mean, respectively. The best results are formatted in bold. }
\begin{center}
\scalebox{0.95}{
\begin{tabular}[t]{ c | c | c | c | c |c |c |c |c  |c |c |c |c |c |c |c | c |c |c }
\specialrule{.1em}{.00em}{.00em}
   \multirow{2}{*}{Methods}   & \multicolumn{3}{c|}{aPaY} & \multicolumn{3}{c|}{AWA1}  & \multicolumn{3}{c|}{AWA2} & \multicolumn{3}{c|}{CUB} & \multicolumn{3}{c|}{SUN} & \multicolumn{3}{c}{FLO} \\ \cline{2-19}  
   
       & U & S & H  & U & S & H  & U & S & H & U & S & H & U & S & H & U & S & H  \\
   
  \hline
\specialrule{.1em}{.00em}{.00em}
  DAP       & 4.8            & 78.3          & 9.0        & 0.0       & \textbf{88.7}    & 0.0   
            & 0.0            & 84.7          & 0.0        & 1.7       & 67.9             & 3.3         
            & 4.2            & 25.1          & 7.2        & -         & -                &-                    
            \\
  CONSE     & 0.0            & \textbf{91.2} & 0.0        & 0.4       & 88.6             & 0.8
            & 0.5            & \textbf{90.6} & 1.0        & 1.6       & 72.2             & 3.1              
            & 6.8            & 39.9          & 11.6       & -         & -                & -                
            \\
  SSE       & 0.2            & 78.9          & 0.4        & 7.0       & 80.5             & 12.9    
            & 8.1            & 82.5          & 14.8       & 8.5       & 46.9             & 14.4    
            & 2.1            & 36.4          & 4.0        & -         & -                & -   
            \\
  SJE       & 3.7            & 55.7          & 6.9        & 11.3      & 74.6             & 19.6
            & 8.0            & 73.9          & 14.4       & 23.5      & 59.2             & 33.6    
            & 14.7           & 30.5          & 19.8       & 13.9      & 47.6             & 21.5
            \\
  LATEM     & 0.1            & 73.0          & 0.2        & 7.3       & 71.7             & 13.3   
            & 11.5           & 77.3          & 20.0       & 15.2      & 57.3             & 24.0    
            & 14.7           & 28.8          & 19.5       & 6.6       & 47.6             & 11.5                      
            \\
  ESZSL     & 2.4            & 70.1          & 4.6        & 6.6       & 75.6             & 12.1    
            & 5.9            & 77.8          & 11.0       & 12.6      & 63.8             & 21.0    
            & 11.0           & 27.9          & 15.8       & 11.4      & 56.8             & 19.0                        
            \\
  ALE       & 4.6            & 73.7          & 8.7        & 16.8      & 76.1             & 27.5    
            & 14.0           & 81.8          & 23.9       & 23.7      & 62.8             & 34.4    
            & 21.8           & 33.1          & 26.3       & 13.3      & 61.6             & 21.9                        
            \\
  SYNC      & 7.4            & 66.3          & 13.3       & 8.9       & 87.3             & 16.2    
            & 10.0           & 90.5          & 18.0       & 11.5      &\textbf{70.9}     & 19.8
            & 7.9            & 43.3          & 13.4       & -         & -                & -                           
            \\
  SAE       & 0.4            & 80.9          & 0.9        & 1.8       & 77.1             & 3.5     
            & 1.1            & 82.2          & 2.2        & 7.8       & 54.0             & 13.6    
            & 8.8            & 18.0          & 11.8       & -         & -                & -                                   \\
  DEM       & 11.1           & 75.1          & 19.4       & 32.8      & 84.7            & 47.3    
            & 30.5           & 86.4          & 45.1       & 19.6      & 57.9            & 29.2    
            & 20.5           & 34.3          & 25.6       & -         & -               & -                            
            \\
  GAZSL     & 14.2           & 78.6          & 24.0       & 29.6	  & 84.2            & 43.8    
            & 35.4           & 86.9          & 50.3       & 31.7	  & 61.3	        & 41.8    
            & 22.1	         & 39.3          & 28.3	      & 28.1      & 77.4            & 41.2
            \\
  GDAN      & 30.4           & 75.0          & 43.4       & -         & -               & -       
            & 32.1           & 67.5          & 43.5       & 39.3      & 66.7            & 49.5    
            & 38.1           & \textbf{89.9} & \textbf{53.4}& -       & -               & -    
            \\
  CADA-VAE  & 31.7              & 55.1             & 40.3          & 57.3      & 72.8            & 64.1    
                & \textbf{55.8}           & 75.0          & 63.9       & \textbf{51.6}      & 53.5            & 52.4    
            & 43.1           & 35.4          & 38.9       & 51.6         & 75.6               & 61.3    
            \\                   
  \hline

  ours      & \textbf{35.0}   & 62.7          & \textbf{44.9}& \textbf{60.4} & 70.4      & \textbf{65.1}
           & 55.2   & 78.9          & \textbf{64.9}& 50.8 & 55.1      & \textbf{52.9}
           & \textbf{44.1}   & 36.8          & 40.1         & \textbf{54.0} & \textbf{79.0}     & \textbf{64.1}
           \\

\specialrule{.1em}{.00em}{.00em}
\end{tabular}}
\end{center}
\end {table*}

\section{Experiments}
\subsection{Datasets and Compared Methods}
To demonstrate the robustness of our method, we conduct experiments on six benchmark ZSL datasets, including three coarse-grained datasets (aPaY \cite{farhadi2009describing}, AWA1\cite{lampert2013attribute}, AWA2\cite{xian2018zero}) and three fine-grained medium-sized datasets (CUB\cite{akata2013label}, SUN\cite{patterson2012sun} and FLO\cite{nilsback2008automated}). The dataset statistics are reported in Table 1. The image features and class embeddings used are provided by \cite{xian2018zero} and are publicly available.
Caltech-UCSD Birds-200-2011 (\textbf{CUB}) consists of 11,788 images from 200 different bird species annotated with 312 attributes. We follow previous zero-shot split in this dataset with 150 classes as seen and 50 as novel classes.

Attribute Pascal and Yahoo (\textbf{aPY}) consists of 18,627 images from the 42 classes and annotated with 64 attributes. It combines datasets a-Pascal and a-Yahoo together, with 30 and 12 classes respectively. Images in a-Yahoo dataset is selected from the Yahoo image search engine to supplement a-Pascal dataset, and the chosen objects are very similar to a-Pascal, e.g. "wolf" vs. "dog".

Animals with Attributes1 (\textbf{AWA1}) is a larger course-grained datasets than aPY in terms of the scale of image number, with 30,475 images of 50 animal species. Each species in the dataset is annotated with 85 attributes, and there are 40 species categorized as seen classes, and 10 as unseen classes.

Animals with Attributes1 (\textbf{AWA2}) collected 37,322 images for the 50 classes of AWA1 dataset from public web sources, e.g. Flickr, Wikipedia, etc. The split scheme and attribute dimension of AWA2 is same as AWA1.

SUN attributes (\textbf{SUN}) is built on top of SUN categorical database to ensure diversity of categories and study the relationship between attribute-based and category-based representations. There are 14,340 images spans 717 categories, in which 645 are split as seen categories and 72 as novel classes. Each category is annotated with 102 attributes.

Oxford Flowers (\textbf{FLO}) includes 8,189 images from 102 flowers categories. The image number of each class is not fixed, ranging from 40 to 258. Also, the images vary in pose, scale and light variation. Since there is no annotated attributes availble for this dataset, we use 2048-dimensional features extracted from semantic descriptions with the RNN model \cite{reed2016learning}.

The dataset statistics is reported in Table 1. We use publicly available image features and class embeddings provided by \cite{xian2018zero}.

We evaluated performance against representative methods proposed over the last few years as well as recent state-of-the-art frameworks. These include DAP \cite{lampert2013attribute}, CONSE \cite{norouzi2013zero}, SSE \cite{zhang2015zero}, SJE \cite{akata2015evaluation}, LATEM \cite{xian2016latent}, ESZSL \cite{romera2015embarrassingly}, ALE \cite{akata2015label}, SYNC \cite{changpinyo2016synthesized}, SAE \cite{kodirov2017semantic}, DEM \cite{zhang2017learning}, 
GAZSL \cite{zhu2018generative}, GDAN \cite{huang2019generative}, CADA-VAE \cite{schonfeld2019generalized}. Of these methods, GAZSL and GDAN learn to synthesise artificial visual representations, then convert the ZSL task into a conventional supervised classification problem. CADA-VAE uses a common embedding space between the visual and semantic modalities to conduct supervised classification. The generative architecture of this model is very similar to our own but does not include the metric learning technique or the uncertainty calibrator mechanism. The classic methods DAP and CONSE first predict the attributes of unseen objects, then infer the attribute labels according to similarity. SSE, SJE, LATEM, ESZSL and ALE all adopt a linear compatibility function or some form of similarity-based metric to compare the embedded visual and semantic features. SAE uses a semantic autoencoder to learn two mappings from the semantic to the visual space, and vice versa from the visual to the semantic space. DAP, ESZSL and SAE perform well on conventional ZSL problems but degrade dramatically when migrating to the GZSL paradigm.

\subsection{Implementation Details}
Our framework is implemented with Pytorch\footnote{http://pytorch.org}. The encoders and decoders are implementated as two-layer fully-connected (FC) networks. The visual, semantic encoders and visual, semantic decoders have 1560, 1450, 1660 and 665 hidden units, respectively. The dimensionality of the latent space is set as 64. All experiments are conducted on a server with 16 Intel(R) Xeon(R) Gold 5122 CPUs and 2 GeForce RTX 2080 Ti GPUs

\subsection{Evaluation Protocol}
The metric used to evaluate both the generalized and conventional ZSL tasks was the widely-used average per-class top-1 accuracy, defined as follows:
\begin{equation}
\begin{aligned}
  Acc_{\mathcal{Y}} = \frac{1}{\norm{\mathcal{Y}}} \sum^{\norm{\mathcal{Y}}}_{c} \frac{\# \  of \ correct \ predictions \ in \ c}{\# \ of \ samples \ in \ \mathit{c}},
\end{aligned}
\end{equation}
where $\mathcal{Y}$ denotes the total class count, and $\mathit{c}$ is the specific class. It is more difficult to achieve a high average per-class top-1 accuracy than the standard accuracy since the former requires good accuracy across all classes.

We use harmonic mean as the evaluation criteria to calculate the joint accuracy of the source and target domains because our goal is to ensure good performance in both domains, and an arithmetic mean could be high simply due to stellar performance in one domain. The formula used to calculate the harmonic mean $\mathcal{H}$ is provided below:
\begin{equation}
\begin{aligned}
  \mathcal{H} = 2 * (acc_{\mathcal{Y}^{tr}}*acc_{\mathcal{Y}^{ts}})/(acc_{\mathcal{Y}^{tr}}+acc_{\mathcal{Y}^{ts}}).
\end{aligned}
\end{equation}
where $\mathcal{Y}^{ts}$ and $\mathcal{Y}^{tr}$ denote the accuracy of seen and unseen classes, respectively. 

\begin{figure}[t]
\centering
\includegraphics[width=\columnwidth]{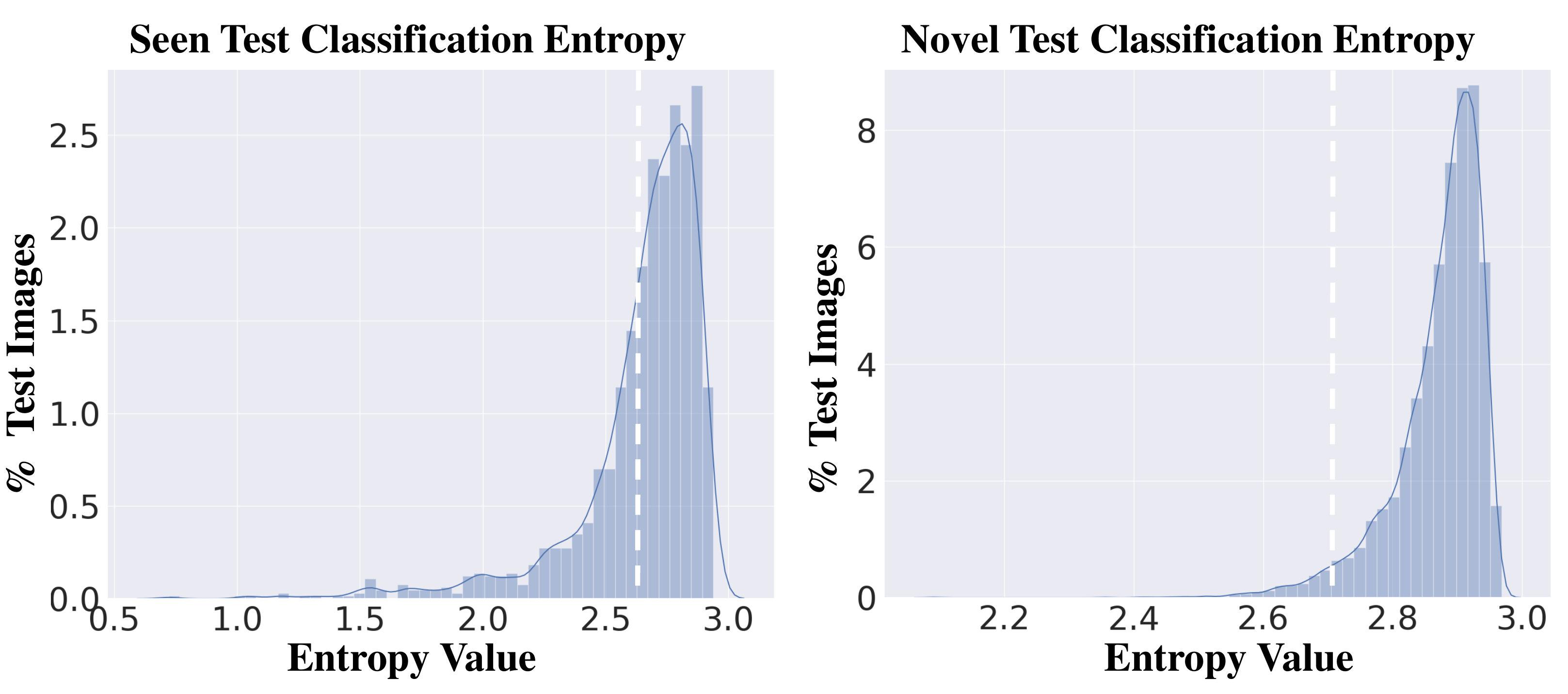} 

\caption{
The probability entropy for the seen and novel test classifications on the aPaY dataset. The white line indicates the entropy threshold (of 2.7) that separates certain seen samples (left) from uncertain ones (right). It is clear that only a few novel test samples were misclassified as seen classes, whereas a large portion of the seen samples were confidently categorized into seen classes.
}
\label{fig1}

\end{figure}

\subsection{GZSL Results}
The results of the GZSL task for all methods are provided in Table 2, showing that our method outperforms the others on most of the datasets. The confusion matrix for the aPaY dataset, which shows classification performance at the class level, is presented in Figure 5. An analysis of these results confirms that the uncertainty calibrator improved model performance with seen objects to a statistically significant degree. At the same time, the results for the unseen classes equal the state-of-the-art. For example, on the aPaY dataset, the visual classifier for the source domain yielded 98\% accuracy on the training set of seen objects, and 91\% accuracy on the test set of seen objects. For the samples certainly from seen classes, the visual classifier is applied to classify them into seen classes. In conventional GZSL settings, seen class accuracy is merely 51.8\% with a general classifier but, with the uncertainty calibrator, it surges dramatically to 62.7\%. The minor trade-off in performance with the novel classes (~ 0.2\%) is entirely worthwhile for an improvement of 41.0\% to 44.9\% in terms of the harmonic mean.

Figure 4 shows the statistics of the classification entropy for the test samples from the aPaY dataset. Notably, the entropy of samples in the seen and novel classes have different distributions. So, when we set the entropy threshold to 2.7, a large amount of the seen test samples fall below the threshold, which means they were confidently classified into the source domain. Conversely, almost all of the novel test samples fall above the threshold are were easily classified into unseen classes. Even with the few misclassifications that slipped through, conventional GZSL approaches struggle in these circumstances as the results show.

\begin{figure}[t]
\centering
\includegraphics[width=0.85\columnwidth]{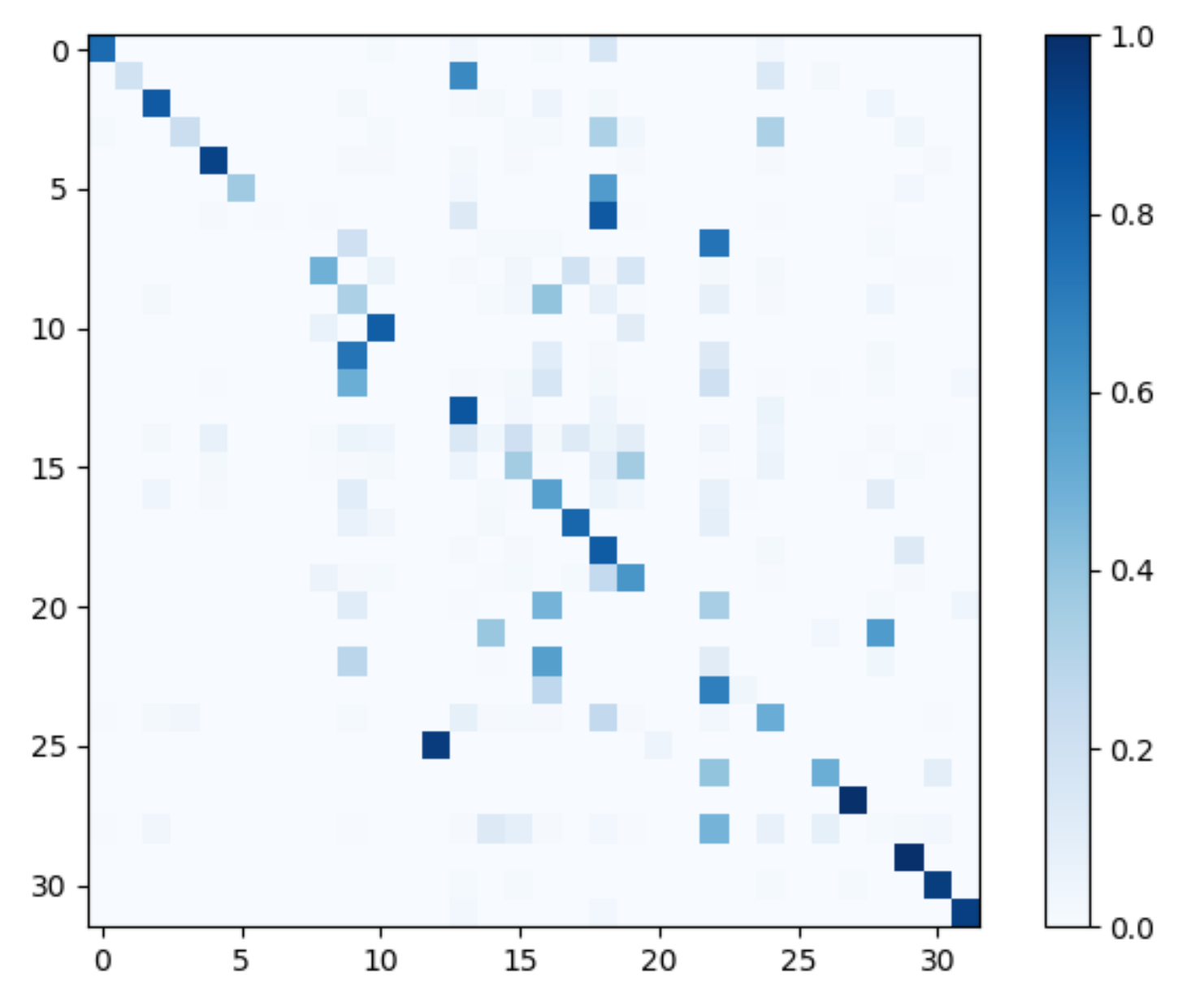} 
\caption{The confusion matrix for the dataset aPaY. The diagonal elements represent the percentage of points for which the predicted label is equal to the true label, while off-diagonal elements were mislabeled by the classifier. }
\label{fig2}

\end{figure}

\begin {table}[t]
\caption {Conventional ZSL accuracy (\%). The best results are formatted in bold.}
\begin{center}
\scalebox{1.1}{
\begin{tabular}[t]{ c | c | c | c | c |c | c}
\specialrule{.1em}{.00em}{.00em}
   & CUB  & aPaY & AWA1 & AWA2 & SUN & FLO \\
  \hline
\specialrule{.1em}{.00em}{.00em}
  DAP               & 40.0  & 33.8     & 44.1           & 46.1    & 39.9    & -   \\
  CONSE             & 34.3  & 26.9     & 45.6           & 44.5    & 38.8    & -   \\
  SSE               & 43.9  & 34.0     & 60.1           & 61.0    &  51.5   & -   \\
  SJE               & 53.9  & 32.9     & 65.6           & 61.9    & 53.7    & 53.4\\
  LATEM             & 49.3  & 35.2     & 55.1           & 55.8    & 55.3    & 60.8\\
  ESZSL             & 53.9  & 38.3     & 58.2           &58.6     & 54.5    & 51.0\\
  ALE               & 54.9  & 39.7     & 59.9           & 62.5    & 58.1    & 48.5\\
  SYNC              & 55.6  & 23.8     & 54.0           & 46.6    & 56.3    & -   \\
  SAE               & 33.3  & 8.3      & 53.0           & 54.1    & 40.3    & -   \\
  GDAN              & 39.3  & 30.4     & -              & 32.1    & 38.1    & -   \\
  DEM               & 51.7  & 35.0     & \textbf{68.4}  & 67.1    & 61.9   \\
  GAZSL             & 55.8   & \textbf{41.1}     & 68.2  & \textbf{70.2} & 61.3 & 60.5  \\
  \hline
  ours   & \textbf{61.73} & 39.1 & 65.7 & 66.0 & \textbf{63.5} &\textbf{67.2} \\
\specialrule{.1em}{.00em}{.00em}
\end{tabular}}
\end{center}

\end {table}

\begin{figure*}[t]
\centering
\includegraphics[width=1.0\textwidth]{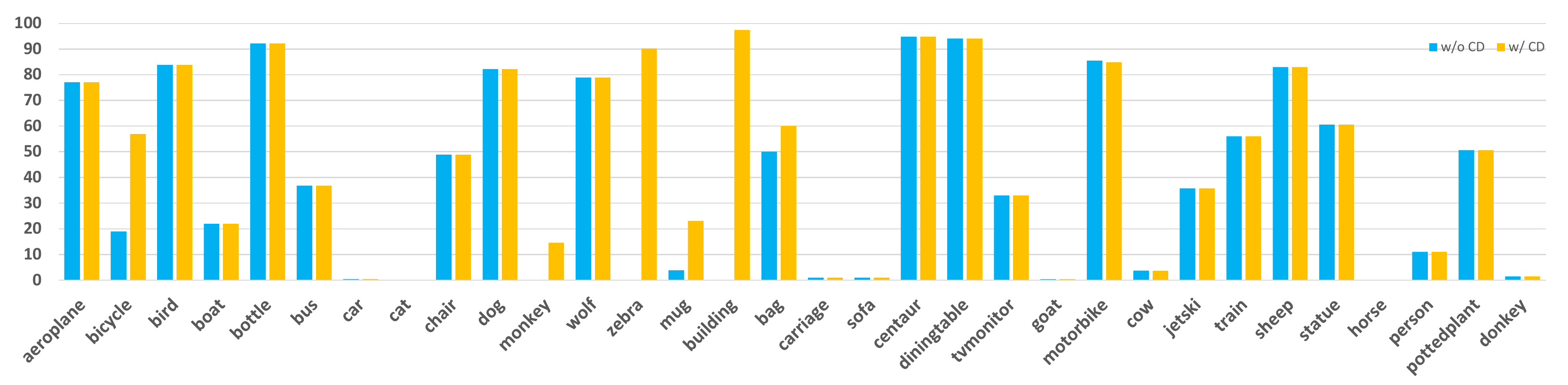} 

\caption{Class accuracy comparison before and after applying uncertainty calibration on dataset aPaY. There are 32 object categories in the dataset, including left 20 categories from source domain and the rest of 12 from novel classes. The blue and yellow histograms illustrate the accuracies for each class without and with the certainty detector, respectively. Most of seen classes witness a major improvement, especially some originally cannot even recognize classes (0 accuracy).}
\label{fig1}
\end{figure*}

\subsection{Conventional ZSL Results}
As a further analysis, we are curious to see how our framework compared with conventional ZSL, where only objects in unseen classes need to be classified. We use our dual VAE model to synthesise a fixed amount of latent representations (400 was optimal in this case), given novel semantic embeddings in n classes, then train an n-way classifier on the supervised data. Again, average per-class top-1 accuracy is the evaluation metric. The results, shown in Table 3, confirm that our approach not only performs well with GZSL tasks but also delivers state-of-the-art performance in conventional ZSL. It is also worth mentioning that the proposed model outperformed all state-of-the-art methods on the CUB, SUN and FLO datasets. The result with the SUN dataset is particularly promising since this dataset has more than 700 categories with only 20 images per category, and accuracy still reached 63.5\%.


\subsection{Ablation Study}
Our ablation study covers the effects of the basic dual VAE framework, the triplet loss and the entropy-based uncertainty calibrator. We train a variant of our model without the entropy-based uncertainty calibrator (EUC), and another without either the uncertainty calibrator or triplet loss and compared performance with the complete framework. The results appear in Table 4. It is clear that each component positively and significantly contributes to performance. The triplet loss improves accuracy by around 2\%, while the uncertainty calibrator improves performance on the seen classes significantly as discussed above.  

\begin{table}[t]
\caption {Ablation study. Effects of different components on GZSL performance (\%) on datasets aPaY and FLO.}
\begin{center}
\scalebox{1.1}{
\begin{tabular}{c|c| c |c |c}

\hline \specialrule{.1em}{.00em}{.00em}
\multicolumn{2}{c|}{Model}                                           & \multicolumn{1}{c|}{\begin{tabular}[c]{@{}c@{}}w/o  \\ EUC \& Trip\end{tabular}} & \multicolumn{1}{c|}{\begin{tabular}[c]{@{}c@{}}w/o  \\ EUC\end{tabular}} & Complete  

\\ \hline  \specialrule{.1em}{.00em}{.00em}

\multicolumn{1}{c}{\multirow{3}{*}{aPaY}}  
& \multicolumn{1}{c}{U} 
& 31.7              & \textbf{35.1}  &      35.0          
\\       \cline{2-5}
\multicolumn{1}{l}{}                      
& \multicolumn{1}{l}{S} 
& 55.1              & 49.2           & \textbf{62.7} 
\\ \cline{2-5}
\multicolumn{1}{l}{}                      & \multicolumn{1}{l}{H} 
& 40.3              & 41.0           & \textbf{44.9} 
\\ \hline

\multicolumn{1}{c}{\multirow{3}{*}{FLO}}  
& \multicolumn{1}{c}{U} 
& 51.6              & \textbf{54.2}  &      54.0         
\\       \cline{2-5}
\multicolumn{1}{l}{}                      
& \multicolumn{1}{l}{S} 
& 75.6              &  77.5          & \textbf{79.0} 
\\ \cline{2-5}
\multicolumn{1}{l}{}                      & \multicolumn{1}{l}{H} 
& 61.3              & 63.8           & \textbf{64.1} 
\\ \hline
\specialrule{.1em}{.00em}{.00em}
\end{tabular}}
\end{center}

\end{table}

\begin{figure*}[t]
\centering
\includegraphics[width=2.1\columnwidth]{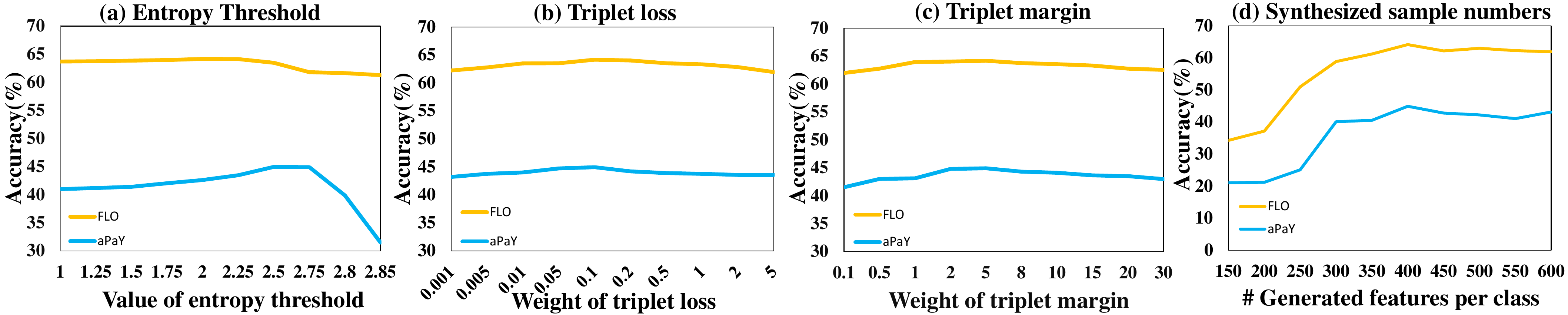} 
\caption{Parameter sensitivity of the proposed method. }
\label{fig1}
\end{figure*}

\subsection{Class-wise Analysis of Calibration}
A comprehensive class-wise accuracy comparison before and after applying the uncertainty calibration is shown in Figure 6. The 20 categories on the left are from the seen classes, while the 12 on the right are the unseen classes. The blue histogram indicates the accuracy for each class without the uncertainty calibrator; the yellow is with. As illustrated, most of the seen classes witnessed a major improvement after applying the entropy-based uncertainty calibration. The object categories that most benefitted are \textit{bicycle}, \textit{monkey}, \textit{zebra}, \textit{mug}, \textit{building} and \textit{bug}. The seen class zebra has the highest boost from an original 0 recognition rate to 90.2\% accuracy. 

\begin{figure}[t]
\centering
\includegraphics[width=1.0\columnwidth]{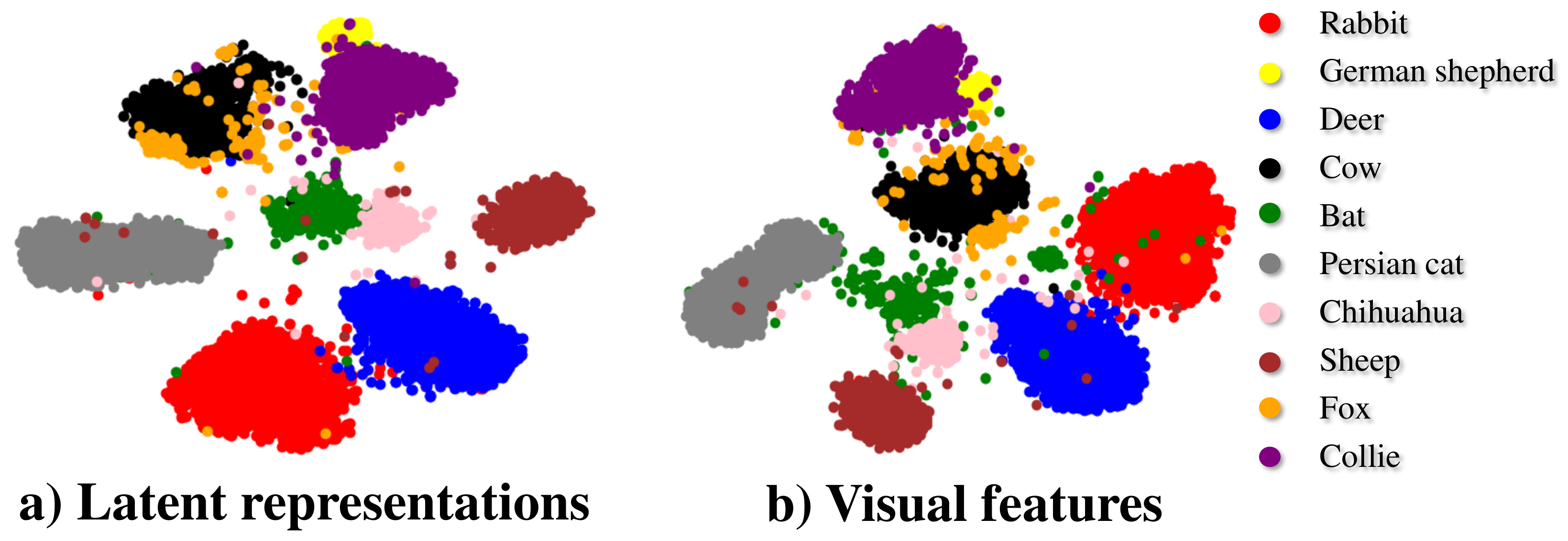} 
\caption{t-SNE visualisation of the distribution of encoded latent features from visual features (left) and visual features (right). It can be seen the encoded visual features are even more discriminative than extracted the visual features.}
\label{fig1}
\end{figure}

\subsection{Parameter sensitivity}
We use the FLO and aPaY datasets to explore the most appropriate hyper-parameter values for the proposed method with GZSL. The entropy-based uncertainty calibrator is, arguably, the most significant component in our overall architecture, and the results in Figure 7(a) indicate that the best thresholds were 2.78 on aPaY and 2.25 on FLO. We observe that performance is very sensitive to this threshold because it controls the percentage of novel samples that are likely to be misclassified into seen classes. Figures 7(b) and 7(c) show the impact of the triplet loss. The optimal weight and margin for the FLO and aPaY datasets is 0.1 and 5, respectively. Further, we vary the number of visual samples generated for training the general classifier, finding 400 samples per class to be the optimal number, as shown in Figure 7(d).

\subsection{Latent Space Distribution Analysis}
To further verify the triplet loss regularisation, we produce a t-SNE visualization with the AWA dataset. Figure 8 visualizes the unseen visual samples from the test set and the corresponding latent features generated by the visual encoder. The distributions of the visual features and the encoded latent features are approximate to each other, which proves that the visual encoder preserves the most visual distribution information for the unseen visual samples. Apart from avoiding the information loss problem, the encoded visual features are more discriminative. For example, the visual features of Bat (marked in green) are mixed with Rabbit (marked in red) and Persian cat (marked in gray). This problem is alleviated by the visual feature encoder. It is particularly noticeable in Figure 8(a) that the data points from Bat are more centralized than the original visual features. Moreover, since the negative class samples are forced to separate during training with triplet loss, the distribution is more inter-class discriminative. To compare the latent features from both modalities, we encoded the semantic vectors as rounded dots shown in Figure 9. The latent features generated from visual features are shown as cross marks. Different colours represents different classes. It can be seen that the visual latent features are centered around the semantic latent features of the same classes.

\begin{figure}[t]
\centering
\includegraphics[width=0.85\columnwidth]{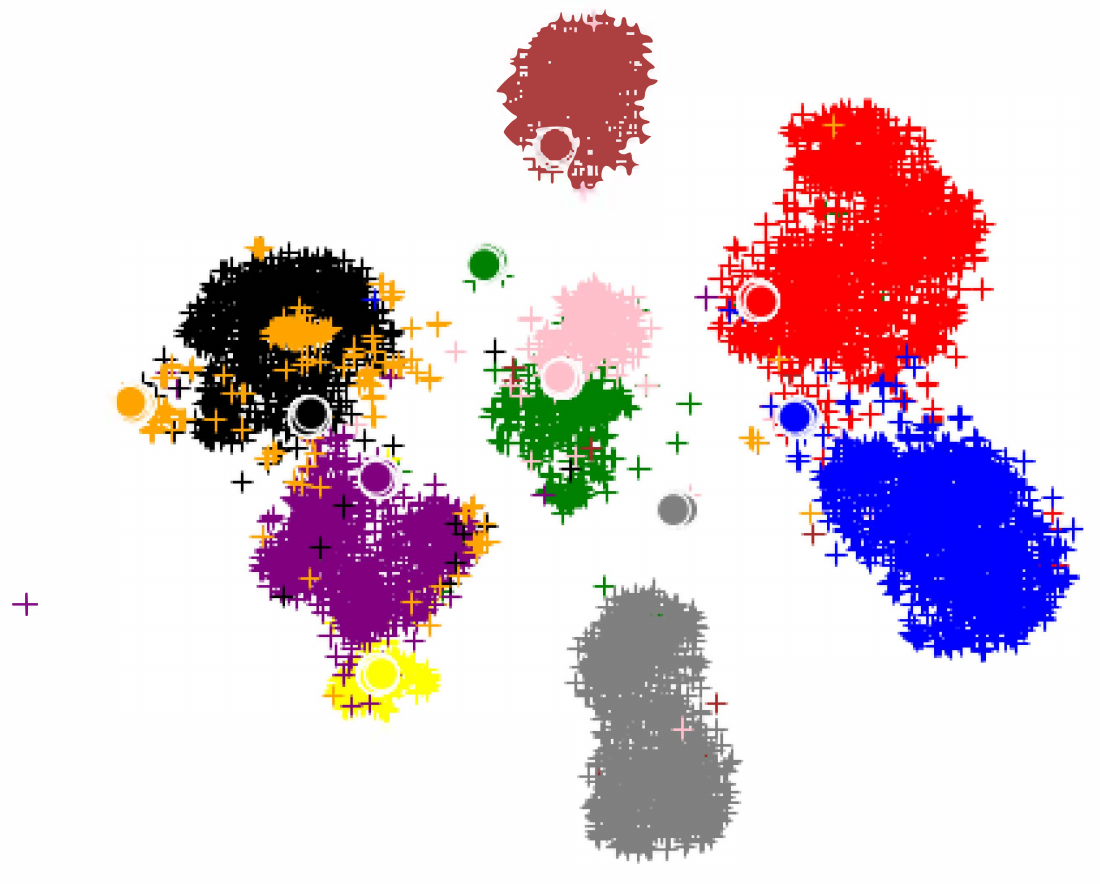} 
\caption{t-SNE visualization of the latent feature distribution of 10 unseen class encoded from visual features ( marked as +) and semantic features (marked as o). Different colours represent different classes.}
\label{fig1}
\end{figure}

\subsection{ZSL from Natural Languages}
Following \cite{zhu2018generative}, we validate our proposed method with natural languages as the semantic information. Considering that annotating attributes is expensive and impractical, we cannot be always annotate attributes for the new categories. Hence, directly using the existing Wikipedia articles as the semantic information is much more practical. Elhoseiny \etal \cite{elhoseiny2017link} collected raw Wikipedia articles for CUB and NAB \cite{van2015building} datasets. They first tokenized the Wikipedia articles into words and got rid of the full stops. In order to weight the terms in the dataset appropriately, Term Frequency-Inverse Document Frequency (TF-IDF) is adopted to extract text feature vectors. The dimensionality of TF-IDF feature for dataset CUB and NAB are 7,551 and 13,217 respectively. There are two splitting designs according to the relationship between seen categories and unseen categories: Super-Category-Shared splitting (SCS) and SuperCategory-Exclusive splitting(SCE). In SCS, unseen categories are chosen to share same super-class with the seen categories. In result, the relevance in this design between seen categories and unseen categories is relatively high. Intuitively, SCE is more difficult to achieve good performance. As for the visual features,  Elhoseiny \etal \cite{elhoseiny2017link} prepared visual features from images in the two bird datasets with Visual Parts CNN Detector/Encoder (VPDE), yielding 3,582 and 3,072 dimensional visual features for CUB and NAB respectively. We compare to other eight state-of-the-art approaches: MCZSL \cite{akata2016multi}, WAC \cite{elhoseiny2013write}, ESZSL \cite{romera2015embarrassingly}, SJE  \cite{akata2015evaluation}, ZSLNS \cite{qiao2016less}, SynC \cite{changpinyo2016synthesized}, ZSLPP \cite{elhoseiny2017link} and GAZSL \cite{zhu2018generative}. In Table \RN{4} We report the performance of our proposed approach and the existing methods, which are the reported from GAZSL\cite{zhu2018generative}. It can be seen from the table that the proposed method consistently outperforms all the counterparts on both datasets and both splits.

\begin {table}[t]
\caption {Top-1 accuracy (\%) on CUB and NAB datasets with two split settings. The semantic information used for these two datasets are the natural languages.}
\begin{center}
\begin{tabular}[t]{ l | c | c | c | c }
\specialrule{.1em}{.00em}{.00em}
   & \multicolumn{2}{c}{CUB}  & \multicolumn{2}{|c}{NAB} \\
  \hline
    Methods              & SCS   & SCE   & SCS   & SCE     \\
\specialrule{.1em}{.00em}{.00em}
  MCZSL\cite{akata2016multi}                 & 34.7  & -     & -     & -       \\
  WAC-Linear\cite{elhoseiny2013write}            & 27.0  & 5.0   & -     & -       \\
  WAC-Kernel  \cite{elhoseiny2013write}          & 33.5  & 7.7   & 11.4  & 6.0     \\
  ESZSL    \cite{romera2015embarrassingly}             & 28.5  & 7.4   & 24.3  & 6.8     \\
  SJE  \cite{akata2015evaluation}                 & 29.9  & -     & -     & -       \\
  ZSLNS \cite{qiao2016less}                 & 29.1  & 7.3   & 24.5  & 6.8     \\
  $SynC_{\textit{fast}}$ \cite{changpinyo2016synthesized}  & 28.0  & 8.6   & 18.4  & 3.8     \\
   $SynC_{\textit{OVO}}$ \cite{changpinyo2016synthesized}  & 12.5  & 5.9   & -     & -       \\
  ZSLPP\cite{elhoseiny2017link}                 & 37.2  & 9.7   & 30.3  & 8.1     \\
  GAZSL\cite{zhu2018generative}                 & 43.7  & 10.3  & 35.6  & 8.6     \\
  \hline
  ours                & \textbf{45.3}  &   \textbf{11.5}   & \textbf{36.7}     &   \textbf{9.7}    \\
\specialrule{.1em}{.00em}{.00em}
\end{tabular}

\end{center}
\end {table}

\subsection{Zero-shot Retrieval}
We also conduct zero-shot image retrieval on CUB and NAB datasets. The same semantic information as zero-shot learning from natural language are leveraged in this task. In zero-shot retrieval, given the semantic information only, the images of unseen classes are to be classified and returned. The commonly used evaluation metric mean average precision (mAP) is used, where the precision is the percentage of correct images retrieved for the class. Table \RN{6} reports the accuracy of each method when retrieving 100\%, 50\% and 25\% of the images from all unseen classes. The retrieval process begins by generating a fixed amount of latent features for the given semantic vectors. We then calculated the mean point among the generated latent features. In the latent space, all the images are embedded as the latent features. We search for the image points that are nearest to the mean point. Since all the images in the datasets are embedded as the latent features with a much lower dimensionality than visual features, the retrieval complexity is relatively lower.

\begin {table}[t]
\caption {Comparison of Zero-shot retrieval on the CUB and NAB datasets (mAP (\%)).}
\begin{center}
\begin{tabular}[t]{  l | c | c | c | c | c | c }
\hline
   & \multicolumn{3}{c|}{CUB}  & \multicolumn{3}{c}{NAB} \\
  \hline
  Methods   & 25\%   &  50\%   & 100\%  &  25\%   &  50\%   & 100\%     \\
\hline 
ESZSL       & 27.9      & 27.3      & 22.7      & 28.9      & 27.8      & 20.9      \\
ZSLNS       & 29.2      & 29.5      & 23.9      & 28.8      & 27.3      & 22.1  \\
ZSLPP       & 42.3      & 42.0      & 36.3      & 36.9      & 35.7      & 31.3     \\
GAZSL       & 49.7      & 48.3      & 40.3      & 41.6      & 37.8      & 31.0     \\
  \hline
ours &\textbf{54.2}  &\textbf{52.6}& \textbf{46.1}  & \textbf{44.5} & \textbf{38.1} & \textbf{33.6}    \\
\hline
\end{tabular}
\end{center}
\end {table}

\subsection{Zero-shot Video Classification}
 Zero-shot learning in video classification is a promising research direction. Our proposed method is also applicable to Zero-shot video classification. We validate our proposed method on video classification under conventional ZSL and GZSL settings. Following \cite{zhang2018visual}, we use datasets HMDB51 and UCF 101, which includes 6,700 and 13,000 videos respectively with 51 and 101 categories. we extract spatio-temporally pooled 4096-d I3D features from pre-trained RGB and Flow I3D networks and concatenate them to obtain 8192-d video features \cite{carreira2017quo}. As for the semantic information, in HMDB51, a skip-gram model \cite{mikolov2013distributed} is used to generate semantic embeddings of size 300, using action class names as input. In UCF101, we use semantic embeddings of size 115, provided with the dataset. We compare to existing state-of-the-art methods for zero-shot video classification in Table \RN{7}. They are HAA \cite{liu2011recognizing}, ST \cite{xu2015semantic}, TZWE \cite{xu2017transductive}, Bi-dir \cite{wang2017zero}, SJE \cite{akata2015evaluation}, ConSE \cite{norouzi2013zero} and GGM \cite{mishra2018generative}. we report the performance comparison to these approaches in Table \RN{7}.

\begin {table}[t]
\caption {GZSL and ZSL performance comparison (in \%) with existing approaches. Best results for each embedding are in bold. }
\begin{center}
\begin{tabular}[t]{ l | c | c  |c |c}
\specialrule{.1em}{.00em}{.00em}
   & \multicolumn{2}{c|}{GZSL}  & \multicolumn{2}{c}{ZSL}    \\
  \hline
    Methods    & HMDB51   & UCF101    & HMDB51   & UCF101     \\
\specialrule{.1em}{.00em}{.00em}
  HAA \cite{liu2011recognizing}             & -  & 18.7 & -    & -       \\
  ST \cite{xu2015semantic}                  & -  & -    & 15.0 & 15.8 \\
  TZWE \cite{xu2017transductive}            & - & - & 19.1 & 18.0 \\
  Bi-dir \cite{wang2017zero}                & - & - & 18.9 & 21.4 \\
  SJE \cite{akata2015evaluation}              & 10.5  & 8.9   & -    & -   \\
  ConSE  \cite{norouzi2013zero}          & 15.4  & 12.7  & -    & -     \\
  GGM    \cite{mishra2018generative}    & 20.1  & 17.5  & 20.7 & 20.3   \\
     \hline
  ours                & \textbf{32.7}  &   \textbf{29.9} & \textbf{28.6} & \textbf{31.4}     \\
\specialrule{.1em}{.00em}{.00em}
\end{tabular}

\end{center}
\end {table}

\section{Conclusion}
In this paper, we proposed a novel framework for GZSL that uses dual VAEs with a triplet framework to learn discriminative latent features. An entropy-based uncertainty calibration minimizes overlapping areas between seen and unseen classes that can lead to performance degradation in seen classes by leveraging the entropy of the softmax probability over those seen classes. Consequently, the classes deemed to be seen with high confidence can be used to train a classifier with a supervised model over seen classes. This feature generation framework proved to be effective with both GZSL and conventional ZSL tasks on the aPaY, AWA1, AWA2, CUB, SUN and FLO datasets. Additionally, the proposed framework is verified on zero-shot learning from natural languages, zero-shot learning image retrieval and zero-shot video classification tasks. However, despite the remarkable results achieved by the framework, the important threshold for the uncertainty calibration must be tuned for each individual dataset. Developing a mechanism for automatically tuning this hyper-parameter is a potential direction for future work.


%



\section*{Acknowledgment}
This work was partially supported by ARC DP190101985, ARC DE200101610, Sichuan  Science and Technology Program under Grant (2020YFG0080) and the National Natural Science Foundation of China (No. 62002085).


\ifCLASSOPTIONcaptionsoff
  \newpage
\fi


\bibliographystyle{IEEEtran}
%
\bibliography{bibliography}

%








\end{document}